# Why current rain denoising models fail on CycleGAN created rain images in autonomous driving[†]

Michael Kranl[1, *], Hubert Ramsauer[2], and Bernhard Knapp[1]

[1]Department of Computer Science, Artificial Intelligence Engineering Program, University of Applied Sciences Technikum Wien, Vienna, Austria
[2]Institute for Machine Learning, Johannes Kepler University, Linz, Austria
*ai21m018@technikum-wien.at

## Abstract

One of the main tasks of an autonomous agent in a vehicle is to correctly perceive its environment. Much of the data that needs to be processed is collected by optical sensors such as cameras. Unfortunately, the data collected in this way can be affected by a variety of factors, including environmental influences such as inclement weather conditions (e.g., rain). Such noisy data can cause autonomous agents to take wrong decisions with potentially fatal outcomes. This paper addresses the rain image challenge by two steps: First, rain is artificially added to a set of clear-weather condition images using a Generative Adversarial Network (GAN). This yields good/bad weather image pairs for training de-raining models. This artificial generation of rain images is sufficiently realistic as in 7 out of 10 cases, human test subjects believed the generated rain images to be real. In a second step, this paired good/bad weather image data is used to train two rain denoising models, one based primarily on a Convolutional Neural Network (CNN) and the other using a Vision Transformer. This rain de-noising step showed limited performance as the quality gain was only about 15%. This lack of performance on realistic rain images as used in our study is likely due to current rain de-noising models being developed for simplistic rain overlay data. Our study shows that there is ample space for improvement of de-raining models in autonomous driving.

## Introduction

Significant progress has been made in the development of autonomous vehicles over the last decade [1]. Much of this progress is due to the availability of increasingly powerful AI systems and models. These systems can make plausible decisions based on the data they collect, just as a human driver is constantly asked to do when driving. One of the core tasks of such an autonomous agent is therefore the correct perception of its environment. A large part of the data required to correctly control and navigate an autonomous vehicle is image data. Optical data is essential for basic tasks such as lane detection, object recognition, distance measurement or calculation of approach speeds [2]. The impeccable quality of this image data is therefore crucial to the smooth functioning of the many processing steps and inference tasks required for autonomous driving.

Unfortunately, optical sensors - mainly cameras - are also subject to a wide range of interference when collecting this data. On the one hand, such interference is caused by the capturing sensors themselves. Every camera sensor produces a certain amount of noise. However, this is due to the way current image sensors work. On the other hand, and with a much more serious impact, such image disturbances are caused by external influences from the environment. This includes adverse weather conditions, such as rain, for example.

This work focuses on the task of image enhancement and restoration using deep learning methods. To achieve this, two essential conditions must be met. First, appropriate training data must be available. Training data must represent a scene with and without the image disturbance. Second, suitable deep learning models capable of removing such disturbances are required. These models must be able to infer from the disturbed image material what the original image would have looked like without the disturbance. This work takes a new approach to the generation of training data, one that has found limited attention in the current literature.

Unlike previous work, the rainy version of a given image pair is not created by applying simple raining layers. Instead, a previously trained Generative Adversarial Network (GAN) is used to perform a style transfer on the training images. This makes it possible to create more realistic rain scenarios. This new type of training data is used to train and benchmark existing de-raining models. Their performance is re-evaluated in the light of this new type of training data.

---

[†] Based on the Master's thesis by Michael Kranl available at UAS Technikum Wien

# Methods

## Training data generation

Deep Learning (DL) models for image restoration and enhancement rely on image pairs during the training process. These image pairs must contain both the clean version of an image without any existing artefacts and a variant containing the artefacts that the model is supposed to remove. Creating a dataset of real images for this type of image interference would be a very challenging task. It would require capturing identical road traffic scenes in both clear and rainy weather. The scene itself, the perspective and the time of day would all have to be the identical. In order to achieve the goal of training DL models for rain removal, we therefore decided to use AI-based methods to generate suitable training data.

The model that was used to generate the training data for the de-raining models is called CycleGAN [3]. It uses an approach that replaces pair-based supervision with set-based supervision. To generate the required training data with CycleGAN, the implementation provided by the authors was used. In order to maintain the current version of the implementation at the time this work was done and to ensure traceability, a fork of their repository was created. It can be accessed using the subrepo in [4].

**bdd_rain dataset** Images from the Berkeley Deep Drive (BDD100K) dataset [5] were used to create the training data for CycleGAN. A python script was used to read the annotations in json format. The script is available at [4]. The images were then automatically sorted into rainy and clear. The images for the dataset of this project were taken from the training part of BDD100K. Of the 70,000 images contained there, there are 5,070 rainy images and 37,344 clear images. Although the two classes are unbalanced this does not affect the result as 1000 images from each of the two classes were randomly selected for the final dataset, forming the bdd_rain dataset used to train the CycleGAN model.

**syn_derain dataset** After generating the bdd_rain training dataset using a CycleGAN, the deraining models were created. Again, images from the BDD100K dataset were chosen as the source material. This time they were chosen, from the test part of the dataset. 1000 images with clear weather labelling were randomly selected. With these images, the versions with artificially added rain were generated in an inference run of the trained CycleGAN model.

## Evaluation of the training data

In the context of this work, it was necessary to evaluate images based on a subjective impression without a reference image being available for comparison. Specifically, the quality of the synthetically generated rain in the syn_derain dataset. Human perception is superior to any metric, no matter how sophisticated [11]. However, a single person's judgement is still biased. To avoid such a one-sided evaluation raising doubts about the objectivity of the work, the evaluation was carried out by different people with different backgrounds. A good way to get a larger audience to judge the images of artificial rain is to use a survey. A survey was designed in the form of a simple quiz. The user was given 10 images to look at in the quiz and should select for each of the images whether it is "real" or "fake".

For the quiz[‡], 6 random images were selected from the rain-set of the syn_derain dataset. It should be emphasised that the selection of the images from the dataset was random. No pre-selection of the images took place that could have biased the results of the survey. For these images, "fake" was taken as the correct answer. Rain-labelled images from the BDD100K dataset were also randomly selected for the remaining 4 images. For these pictures, the correct answer was "real". The outcome of the survey es presented in the results section.

## Training and evaluation of de-raining models

The de-raining models used for evaluation in this work are Convolutional Neural Networks (CNNs) and transformer based. As there are not many published models in the transformer sector for this special use case, the choice fell on Restormer [6] which represents the first de-raining model evaluated in the course of this work. The main reason for selecting this model was that the authors made a conscious effort to reduce the computational complexity of the model.

---

[‡] The user survey can be accessed in https://forms.gle/QVXYhwdDfQAbMJtk7

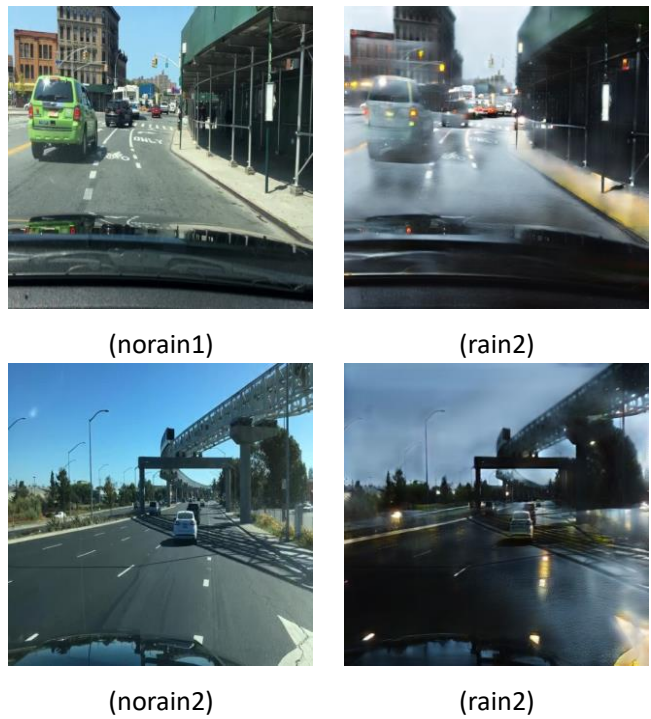

(norain1)   (rain2)

(norain2)   (rain2)

Figure 1: Representative synthetic rain image examples. Left side: original image from the BDD100K dataset. Right side: same image but with rain conditions artificially added by the trained CycleGAN model.

The second model examined in this work is called Recurrent SE Context Aggregation Net (RESCAN) [7]. The model works mainly with CNNs and partly with Recurrent Neural Networks (RNNs).

To train both models, existing implementations provided directly by the authors were used. To perform the training of the Restormer model the implementation found in the Restormer sub-repository in [4] was used. Several changes were required to adapt the existing implementations to the needs of the current project. A relatively profound change was the adjustment of the model's batch size and patch size. This change was required to allow model training with the limited VRAM available in the used hardware setup. A further adjustment had to be made because the original model had only been trained on the Rain100H and Rain100L [8] datasets and therefore only supported the directory structure used there. The implementation was changed to allow any model name to be specified for training. Parameters used for training can be found in the repository mentioned above.

Only minor changes were required in the RESCAN implementation which can be found in the RESCAN subrepo in [4]. However, the structure of the syn_derain dataset had to be adapted for RESCAN. The image pairs do not need separate directories for the rain and norain versions of the images. Instead, the pairs must be merged into one image, which then has the dimensions 2W×H of a single image. In this merged image, the rain version must be on the left and the clean version on the right.

In the validation process, the trained models and the images from the test dataset (*n*=94) of syn_derain were used to create predictions. The predictions were performed on the rainy images from the said test dataset. The result of the predictions was a derained image. The qualitative results compared to the clean initial image are presented in the results section.

## Results

### Training data of paired clear/rain weather images

Using 1000 clear weather images from the BDD100K dataset, artificially rainy versions of these images were created using the trained CycleGAN model for image-to-image translation. This set of image pairs was splitted into a train, test and validation subset. The test subset consisting of 94 image pairs was used later to measure the performance of the de-raining models. Example results from the syn_derain dataset generated in this way are shown in fig. 1.

## User survey

The quiz had a total of 21 participants over a period of about 2 weeks. For the evaluation of the answers, each answer was considered separately and no grouping per picture was made. Thus, 210 independent answers were included in the evaluation. In order to visualise the results a confusion matrix is depicted in fig. 2. The calculated metrics for the survey result can be found in table 1. False Positive Rate (FPR) and accuracy have relatively low values, which shows that the participants had difficulties in deciding whether a synthetic image was real or fake. Taken together this indicates a rather high quality of the generated rain images and shows that the images are well suited for training the de-raining models.

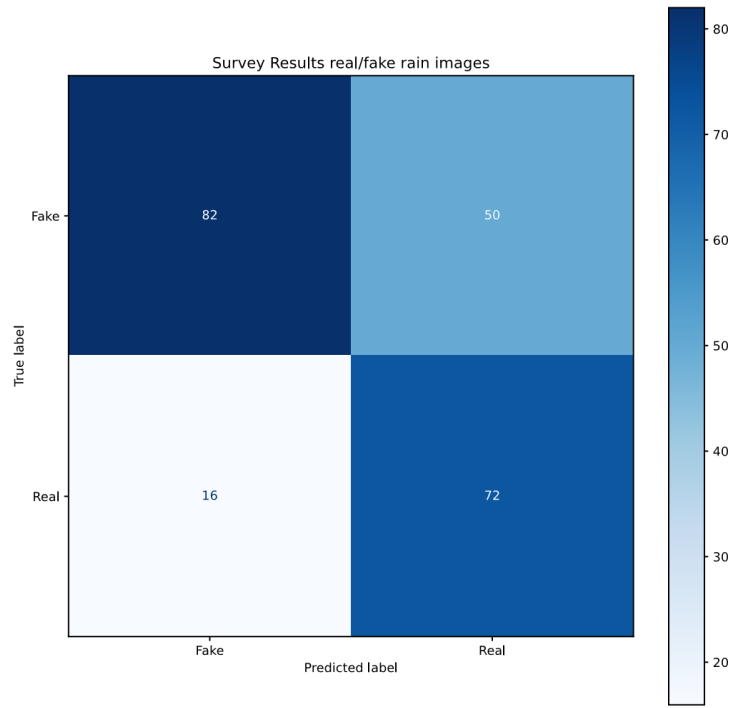

Figure 2: Survey confusion matrix for 210 answers of 21 participants

Table 1: Real/fake survey results

| Metric | Value |
| --- | --- |
| FPR | 37.8% |
| TPR | 81.8% |
| Precision | 59.0% |
| Accuracy | 73.3% |

## De-raining models

The results of the de-raining performance evaluation of Restormer and RESCAN can be found in table 2. Structural SIMilarity (SSIM) [9] and Peak signal-to-noise Ratio Human Visual System Model (PSNR-HVS-M) [10] were used as quality assessment metrics. The table also shows the average quality values of the rained images before the respective de-raining model was applied to them. These values can be found in the row *Rain Image*. These values are used as a reference value to show the improvement achieved over the rained image. The average quality gain compared to the rainy image and broken down by de-raining model and quality index is listed in the *Gain* column. Fig. 3 shows the distribution of the quality scores. Fig. 4 shows some example images of the de-raining model results.

Table 2: Quality results for 94 samples of syn_derain test dataset

|  | SSIM | | | PSNR-HVS-M | | |
|---|---|---|---|---|---|---|
|  | Mean | σ | Gain | Mean | σ | Gain |
| Restormer | 0.79 | 0.11 | +0.16 | 20.68dB | 5.50dB | +4.49dB |
| RESCAN | 0.74 | 0.19 | +0.11 | 18.44dB | 1.71dB | +2.25dB |
| Rain image | 0.63 | 0.16 | - | 16.19dB | 3.60dB | - |

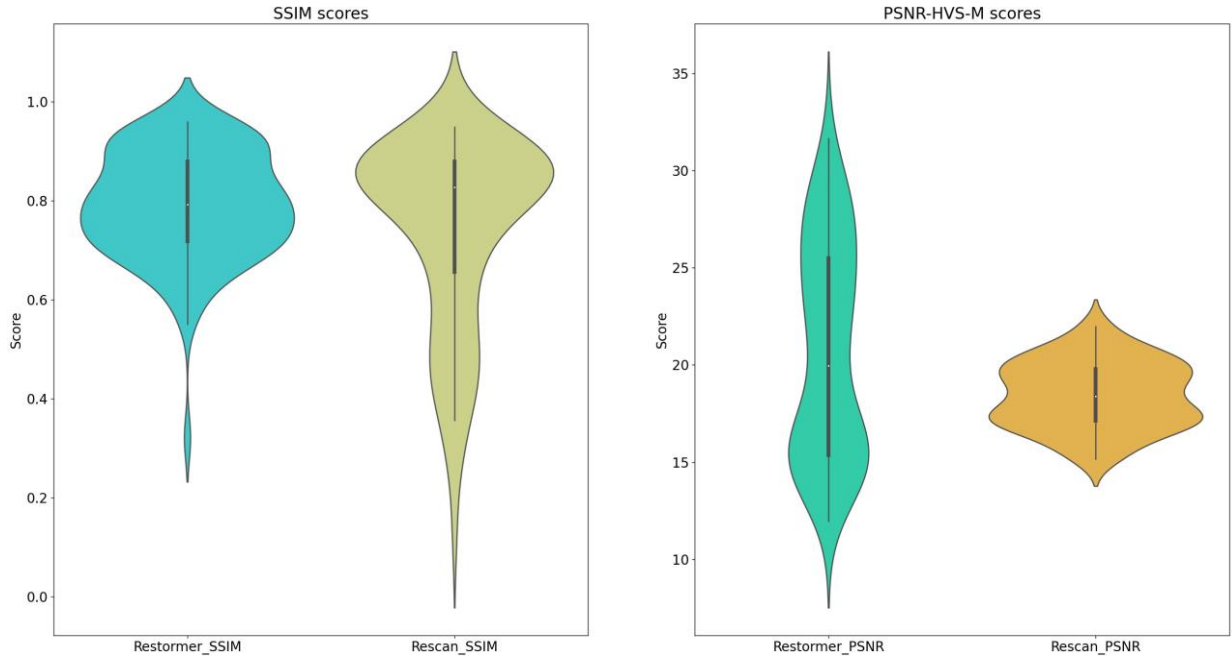

Figure 3: Violin plot of de-rained images (*n*=94) quality scoring

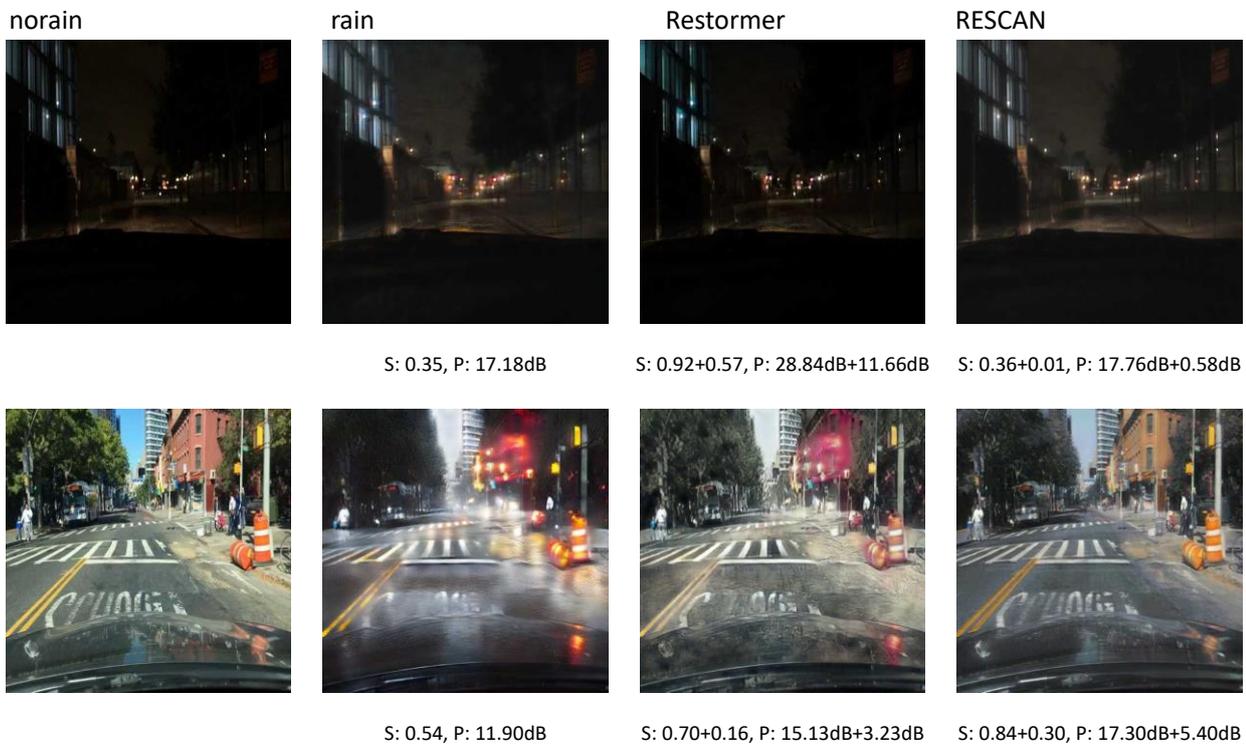

Figure 4: Example results of Restormer and RESCAN Model. To the left is the reference image without rain (*norain* column) and next to it is the version of this image that has been rained using CycleGAN (*rain* column). The third image from the left shows the derained result of Restormer (*Restormer* column) and the right image shows the derained result of RESCAN (*RESCAN* column). Below each image (except the reference image), the *S* value shows the corresponding SSIM quality rating, and the *P* value shows the rating using PSNR-HVS-M. In the Restormer and RESCAN columns, the quality gain over the rain image is also given after the rating. The gain value is positive if the result of the respective de-raining model achieved a better-quality rating compared to the rained image.

## Discussion

Our results show that the generation of clean/rainy image pairs using a cycle GAN is a promising approach. The results of the user survey show that the image-to-image translation works well for rainy weather conditions and is sufficiently realistic for human perception. This shows that it is possible to train de-raining models with realistic data and not just use overlays with artificial rain streaks as usually done in the literature [12, 13]. Reflections from a wet road, for example, can interfere with sensor perception. This can cause object recognition to fail and the autonomous agent to overlook pedestrians or other road users.

Regarding the results of the de-raining models, the quality gains are rather small. The main reason for this is that the de-raining models used are primarily designed to remove artificial rain streak overlays. RESCAN tries to cover a wide range of rain patterns with several streak layers in order to learn as many different patterns as possible. However, as the real rain images from the BDD100K dataset show, such intense rain streaks are rarely encountered in road traffic situations. Furthermore, these streaks are usually not the main cause of visibility impairment but are also caused by reflections from wet road surfaces and spray. A quality comparison with existing work was not useful in the given context. During the research, we did not come across any projects using training data generated by CycleGAN (or other image-to-image translation models) to train image restoration models. Other authors have always used datasets with rather simple rain overlays. The authors of Restormer [6] and RESCAN [7] also used only such datasets.

## Conclusion

Our study shows that current rain noising models have limited performance on realistic rain images. Therefore, we believe that new and/or optimised methods are needed in order to properly remove bad weather influences from images in autonomous driving.

## Data and Software Availability

The used software, models and data sets are available in the following GitHub repository: https://github.com/MicKr4ne/iwire-av.